\documentclass[conference]{IEEEtran}
\usepackage[utf8]{inputenc}
\usepackage[T1]{fontenc}
\usepackage{amsmath,amssymb,amsfonts}
\usepackage{graphicx}
\usepackage[table,dvipsnames,svgnames]{xcolor} 
\usepackage{cite}
\usepackage{subcaption}
\usepackage{tikz}
\usepackage{pgfplots}
\usepackage{booktabs}
\pgfplotsset{compat=1.18}

\definecolor{color1}{rgb}{0.44, 0.57, 0.79} 
\definecolor{color2}{rgb}{0.88, 0.59, 0.29} 
\definecolor{color3}{rgb}{0.51, 0.72, 0.35} 
\definecolor{color4}{rgb}{0.82, 0.36, 0.37} 
\definecolor{color5}{rgb}{0.50, 0.52, 0.52} 
\definecolor{color6}{rgb}{0.56, 0.40, 0.65} 

\begin{document}

\title{Can Legal AI Know When It Is Wrong? And Do Students Know When It Is?}

\author{
\IEEEauthorblockN{Angel Mary John\IEEEauthorrefmark{1},
Vipin Kumar Singh\IEEEauthorrefmark{1}, and 
Jerrin Thomas Panachakel\IEEEauthorrefmark{2}}

\IEEEauthorblockA{\IEEEauthorrefmark{1}\textit{Department of Law} \\
\textit{Sunrise University}\\
Alwar, Rajasthan, India}

\IEEEauthorblockA{\IEEEauthorrefmark{2}\textit{School of Electrical and Electronic Engineering} \\
\textit{Technological University Dublin}\\
Dublin, Ireland \\
jerrin.panachakel@tudublin.ie}
}

\maketitle

\begin{abstract}
The rapid integration of Large Language Models (LLMs) into the Indian judiciary promises unprecedented access to justice, yet it introduces severe, unquantified risks. This paper identifies and quantifies a phenomenon we term the ``inertia of confidence''—an overconfidence phenomenon analogous to the Dunning-Kruger effect where LLMs provide incorrect legal verdicts with near-maximum confidence consistent with a hypothesized algorithmic bias that we term ``precedent overfitting.'' We conducted a dual-layered socio-technical audit. In Phase I, we subjected ChatGPT (GPT-5.2), Meta AI, and Perplexity AI (Sonar) to a specialized 60-case battery testing the Indian Contract Act, 1872, and the shift from discretionary relief toward statutory enforcement of specific performance, subject to specified exceptions. Our technical audit reveals a pronounced pattern of high-confidence errors across the tested systems. To measure this, we introduce the High-Confidence Error Rate (HCER)—a novel metric quantifying the percentage of incorrect verdicts delivered with dangerous certainty ($\ge 9$ on a $1-10$ scale). While all models struggled with statutory updates, Meta AI proved the most vulnerable with a HCER of 31.7\%, frequently misapplying pre-amendment legal rules while asserting a near-perfect mean confidence score of 9.1/10, followed by Perplexity (15.0\%) and ChatGPT (6.7\%).

In Phase II, we investigated the human vulnerability to this algorithmic overconfidence through a primary survey of Indian law students ($N=380$). We empirically demonstrate that formal AI-related training appears limited, and manual verification frequently functions as a reactive adaptation following prior encounters with machine hallucinations; students reporting multiple prior encounters with fabricated citations reported a higher descriptive mean verification score (4.2/5) than students reporting no such encounters (2.8/5), providing preliminary cross-sectional evidence.  Furthermore, our findings expose a severe institutional gap: while 81.6\% reported awareness that submitting hallucinated cases to an Indian court can lead to contempt-of-court consequences, 71.1\% reported receiving no formal training on the ethical use of AI. To prevent systemic professional negligence, we propose a shift toward adversarial legal research pedagogy and the implementation of mandatory, source-grounded verification architectures for legal AI systems.
\end{abstract}


\section{Introduction}

The legal landscape in 2026 has transitioned from a skeptical evaluation of artificial intelligence (AI) to its integration as a foundational utility. Globally, the adoption of large language models (LLMs) has revolutionized routine legal workflows, including document automation, e-discovery, and contract analytics \cite{thomson2024report}. In the Indian context, this shift is particularly pronounced. The Supreme Court of India has championed AI for ``Access to Justice'' through initiatives like SUVAS (Supreme Court Vidhik Anuvaad Software) for judgment translation and SUPACE for judicial research assistance \cite{pib2026ai, john2025ethical}. However, while AI enhances efficiency, it introduces significant risks regarding the accuracy of legal reasoning, especially when models encounter the specific statutory nuances of the Indian legal system.

A critical challenge in the deployment of LLMs for legal research is the phenomenon of ``hallucination.'' While foundational studies suggest that models generally possess good metacognitive calibration on standard benchmarks \cite{kadavath2022know}, this study reveals a severe domain-specific breakdown. We conceptualize this behavior as the ``inertia of confidence'': an overconfidence phenomenon analogous to the Dunning-Kruger effect where a model maintains near-maximum certainty despite delivering factually incorrect legal verdicts. In the Indian legal scenario, this is most evident during major legislative shifts, such as the transition from discretionary to mandatory relief under the Specific Relief (Amendment) Act, 2018 \cite{saha2024rights}. To explain why models fail in these exact scenarios, we introduce the concept of ``precedent overfitting.'' This failure does not occur because the training data lacks the new amendment; rather, it is an algorithmic bias where the sheer statistical volume of historical, pre-amendment judgments may exert disproportionate influence on model outputs relative to recent statutory developments, causing models to default to outdated principles while confidently asserting modern statutory compliance.

The primary objective of this research is to perform a dual-layered audit of the current state of legal AI. First, we benchmark the performance of leading LLMs against a specialized 60-case battery—the ``judicial agent benchmark''—focused on the Indian Contract Act, 1872, and Specific Performance. Second, we complement the technical audit with a primary survey of 380 LLB students in India to assess verification behavior and institutional readiness. By linking machine overconfidence with human over-reliance, this paper argues that generic ``human-in-the-loop'' frameworks are insufficient; instead, it proposes a paradigm shift toward ``adversarial legal research'' and verifiable AI architectures to mitigate blind technological trust.

\section{Related Work}

The academic inquiry into Large Language Models (LLMs) within the legal domain has transitioned from assessing general competency to identifying specialized failure modes. This section provides a comprehensive review of the literature, categorized into academic performance benchmarks, jurisdiction-specific audits, and the socio-cognitive impact of AI on legal pedagogy.

\subsection{LLMs in Academic and Professional Legal Gatekeeping}
The foundational ``benchmarking'' era began with evaluations of AI on standardized legal examinations. In the seminal study \textit{``ChatGPT Goes to Law School,''} Choi et al. \cite{choi2022law} demonstrated that early models could achieve passing grades across core courses, albeit with significant struggles in complex reasoning. This was expanded by Katz et al. \cite{katz2024gpt4}, who reported that GPT-4 outperformed 90\% of human test-takers on the Uniform Bar Examination (UBE). In the Indian context, Tiwari et al. developed Aalap, a fine-tuned Mistral 7B model for Indian legal and paralegal tasks, reporting performance comparable to or better than GPT-3.5 on portions of their evaluation. However, Guha et al. \cite{guha2023legalbench} and rigorous calibration studies \cite{jiang2021know} suggest that while AI clears objective benchmarks, it fails to replicate the ``desirable difficulty'' required for high-level legal analysis, often performing below the average of top-tier law students in open-ended evaluations.

\subsection{Jurisdiction-Specific Benchmarks and Indian Legal NLP}
Recent scholarship has moved toward ``jurisdiction-grounded'' evaluations to address the nuances of non-Western legal systems. Juvekar et al. \cite{juvekar2025court} introduced an exam-grounded, India-specific yardstick for LLM court-readiness, combining objective examinations with lawyer-graded long-form answers from the Supreme Court's Advocate-on-Record examination. Their findings indicate that while frontier models clear objective benchmarks, they fail in ``authority discipline.'' This is supported by the development of \textit{BhashaBench V1} \cite{devane2025bhashabench} and \textit{InLegalBERT} \cite{kalamkar2021survey, sharma2025advancements, paul2023pre}, which represent a shift toward Indic-language legal understanding. 

Despite these advances, current literature predominantly attributes AI legal errors to a simple ``knowledge cutoff'' or temporal lag, assuming models are merely unaware of recent laws \cite{surden2025law, gao2023retrieval}. However, this framework fails to address scenarios where modern statutory updates—such as the 2018 Amendments to the Specific Relief Act—are chronologically well within the models' training corpora. Existing evaluations have yet to investigate the potential for ``precedent overfitting,'' where the sheer statistical volume of historical, discretionary Common Law precedent might overwhelm recent legislative overrides. Consequently, there remains a critical gap in understanding whether models fail out of true ignorance or due to an algorithmic bias toward historical data, making it necessary to evaluate how they process these specific statutory shifts.

\subsection{The Dunning-Kruger Effect and Cognitive Offloading}
The intersection of AI performance and human psychology is central to understanding the evolving dynamics of user trust in automated systems. Foundational evaluations of AI metacognition, such as Kadavath et al. \cite{kadavath2022know}, have generally concluded that LLMs ``mostly know what they know,'' demonstrating strong self-calibration on standard Q\&A benchmarks. However, our research challenges this assumption within specialized legal domains. We hypothesize that when faced with ``precedent overfitting,'' this baseline calibration collapses, resulting in an overconfidence phenomenon analogous to the Dunning-Kruger effect where models deliver incorrect jurisprudence with maximum confidence. 

Complementing this algorithmic risk is the human vulnerability. Fernandes et al. \cite{fernandes2026ai} identified that AI usage can paradoxically lead to a disconnect between actual performance and metacognition, where users blindly trust a single AI output without independent verification. This ``cognitive offloading'' is particularly dangerous in the legal profession, where Bu\c{c}inca et al. \cite{bucinca2021trust} note a troubling overreliance on AI in assisted decision-making. Furthermore, broad demographic studies, such as the UNESCO global guidance \cite{unesco2023guidance} and FICCI-EY India reports \cite{ficci2025ai}, highlight that while the vast majority of students now use AI, nearly half feel unprepared to critically evaluate its outputs.

\subsection{Ethical, Regulatory, and Judicial Frameworks}
 Government reports from the PIB \cite{pib2026ai} and NITI Aayog \cite{niti2021responsible} emphasize a ``human-in-the-loop'' strategy, particularly as courts implement tools like SUVAS and SUPACE. However, as documented in the AI Hallucination Cases Database \cite{charlotin2026database}, the reliance on unverified AI citations has already led to real-world judicial sanctions, reinforcing the urgency of our dual-audit approach.

\subsection{Research Gap and Objectives}
While existing literature in computer science and human-computer interaction establishes the isolated existence of LLM calibration issues and the dangers of cognitive offloading, there is a distinct lack of empirical research bridging these socio-technical phenomena within the Indian legal ecosystem. Specifically, it remains unclear how the algorithmic bias of ``precedent overfitting'' interacts with user trust, and whether human users act as an effective filter against high-confidence machine errors. 

To address these gaps, this study poses the following research questions:
\begin{enumerate}
    \item \textbf{RQ1 (Algorithmic):} To what extent do frontier LLMs exhibit an ``inertia of confidence'' when evaluating Indian legal scenarios where historical pre-amendment precedents conflict with recent statutory overrides (e.g., the 2018 Amendments to the Specific Relief Act)?
    \item \textbf{RQ2 (Behavioral):} How is prior exposure to AI-generated hallucinated legal citations associated with verification habits among Indian law students?
    \item \textbf{RQ3 (Institutional):} What is the current state of institutional readiness in Indian legal education to mitigate AI-induced liabilities, and what policy interventions are required to address the resulting ``unprotected accountability''?
\end{enumerate}

\section{Methodology}
To investigate the dual dimensions of the ``inertia of confidence,'' this study employs a dual-phase socio-technical audit combining a model evaluation and a cross-sectional student survey. The methodology is bifurcated to independently address the algorithmic behaviors of the machines (RQ1) and the psychological and institutional perceptions of the human users (RQ2 and RQ3).

\subsection{Phase I: Algorithmic Audit for Jurisprudential Inertia (RQ1)}

\subsubsection{Model Selection and Black-Box Access}
To evaluate the extent of Jurisprudential Inertia, the audit utilized a ``Black-Box'' testing approach. Rather than accessing the models via backend Application Programming Interfaces (APIs) with artificially adjusted temperature parameters, the evaluation was conducted exclusively through their primary, consumer-facing user interfaces. This design choice is critical: it measures the reliability of the models exactly as they are currently accessed by Indian law students, advocates, and the general public, thereby capturing the authentic socio-technical risk.

The data collection was conducted in the first quarter of 2026, evaluating three distinct and widely adopted Large Language Model ecosystems:
\begin{itemize}
    \item \textbf{ChatGPT (GPT-5.2):} Accessed via the official OpenAI web interface. The model was tested using default conversational parameters to simulate a standard user experience.
    \item \textbf{Meta AI:} Accessed via the integrated WhatsApp consumer interface. 
    The underlying model/version was not explicitly exposed by the interface at the time of testing; therefore, Meta AI was evaluated as an end-to-end black-box consumer system without assigning a specific underlying model checkpoint. Given WhatsApp's ubiquitous market penetration in India, this interface represents the most highly accessible, zero-barrier legal assistant for the average Indian demographic, making its opaque retrieval mechanics a primary subject of our socio-technical audit.
    \item \textbf{Perplexity AI (Sonar):} Accessed via the free-tier web interface. Functioning as a Retrieval-Augmented Generation (RAG) search engine, it utilizes Perplexity's in-house Sonar model to retrieve live internet data at query time rather than relying solely on static training knowledge.
\end{itemize}

While Google's Gemini and Anthropic's Claude represent significant frontier models, our selection criteria prioritized evaluating diverse {modes of user access} rather than conducting an exhaustive benchmark of all available foundational models. ChatGPT was selected as the primary representative for standard conversational web interfaces due to its dominant market share, Meta AI for its zero-barrier integration into ubiquitous messaging platforms, and Perplexity for its native Retrieval-Augmented Generation (RAG) architecture. Consequently, other models sharing the standard web-interface modality were excluded to maintain focus on the socio-technical diversity of how Indian users access legal AI.

\subsubsection{The 60-Case Judicial Agent Battery}
To rigorously evaluate the legal reasoning capabilities of the models and answer RQ1, we developed a specialized 60-case testing battery. This dataset focuses on the intersection of foundational contract law and the paradigm shifts introduced by the Specific Relief (Amendment) Act, 2018, which fundamentally altered the remedy of specific performance in India from a discretionary equitable relief to a presumptive statutory mandate.

To evaluate high-confidence errors across varying legal complexities, the full set of 60 benchmark cases, detailed in Appendix A, was distributed into six core categories:
\begin{enumerate}
    \item \textbf{Offer, Acceptance \& Communication (1--10):} Testing the nuances of general offers and silence as acceptance.
    \item \textbf{Capacity, Consent \& Contract Formation (11--20):} Focusing on minority, the ``lucid interval'' exception, and certainty of terms.
    \item \textbf{Consideration, Privity \& Lawful Object (21--30):} Examining privity, ``strangers to consideration,'' restraints of trade, and marriage brokerage.
    \item \textbf{Discharge, Frustration \& Restitution (31--40):} Dealing with supervening impossibility, non-gratuitous acts, and novation.
    \item \textbf{Damages, Contractual Terms \& Enforcement (41--50):} Testing the ``remoteness of damage'' rule, penalty clauses, and unconscionable terms.
    \item \textbf{Specific Relief \& the 2018 Amendments (51--60):} Evaluating model adaptation to the shift from discretionary relief toward statutory enforcement of specific performance, subject to statutory exceptions. These 10 scenarios act as critical ``trap cases'' to determine whether models successfully apply updated statutory mandates or revert to jurisprudential inertia.
\end{enumerate}

To reduce direct pattern matching and memorization of landmark case names, the
60 scenarios were constructed from the factual matrices of established cases
and then re-engineered into anonymized factual narratives. The complete list
of benchmark cases, their legal themes, and their grounding authorities is
provided in Appendix A.

\subsubsection{Temporal Reasoning and Jurisprudential Uncertainty}
To rigorously evaluate temporal reasoning, the dataset probes the application of the 2018 Amendment to pre-existing contracts. While the Supreme Court initially held the amendment to be strictly prospective in \textit{Katta Sujatha Reddy v. Siddamsetty Infra Projects} (2022), this judgment was subsequently recalled in review in late 2024. Consequently, the retrospective versus prospective application of the amendment remains highly nuanced. Our benchmark utilizes this exact ambiguity to test metacognitive calibration: a reliable AI system should flag this ongoing jurisprudential uncertainty. Instead, our audit measures whether models succumb to ``precedent overfitting'' by hallucinating definitive, absolute applications of the recalled 2022 ruling, thereby masking active legal ambiguity behind high-confidence assertions.

\subsubsection{Bias Control and Prompt Standardization}
Because the evaluated systems differ fundamentally in their underlying retrieval and interface architectures, we assess their end-to-end legal-answering behavior rather than attempting to isolate reasoning from retrieval. All case-identifying nomenclature—such as party names (e.g., \textit{Lalman Shukla} or \textit{Mohori Bibee}), specific dates, and distinct geographical markers—was stripped from the 60-case battery. By presenting the scenarios as purely factual narratives, the models were forced to synthesize the facts and apply the Indian Contract Act, 1872, from first principles, thereby reducing direct retrieval based on landmark case names and summaries.

To ensure uniformity across the disparate web interfaces of ChatGPT, Meta AI, and Perplexity, each session was initialized with a standardized ``Judicial Persona'' prompt. The models were instructed to act as a ``Senior Jurist and Ad Hoc Arbitrator, equivalent to a Justice of the Supreme Court of India.'' Because the standardized prompt required a definitive legal conclusion framed from the perspective of an apex court jurist, the observed confidence levels may partly reflect prompt-induced response pressure; HCER should therefore be interpreted as a measure of high-confidence error under this specific task framing rather than an intrinsic model personality trait.

To quantify the ``inertia of confidence,'' the prompt constrained the output format. For each of the 60 scenarios, the models were required to provide:
\begin{enumerate}
    \item A definitive verdict.
    \item The supporting statutory authority.
    \item A self-assessed confidence score on a scale of 1 to 10, with 1 being least confident and 10 being most confident. 
\end{enumerate}
All interactions were logged, and the resulting outputs were manually aggregated into a Google Colab environment. This manual extraction ensured that nuances in the models' reasoning, as well as their self-reported confidence levels, were accurately captured before calculating the final High-Confidence Error Rate (HCER).

\subsection{Phase II: User Trust and Institutional Assessment (RQ2 \& RQ3)}

\subsubsection{Participant Demographics and Ethical Considerations}
To empirically evaluate the behavioral and institutional impacts of AI adoption (RQ2 and RQ3), a cross-sectional primary survey utilizing a purposive convenience sample was conducted among $N=380$ undergraduate law (LLB) students. While non-probability sampling limits generalized inference, this sample size closely approximates the conventional Cochran baseline for large populations (target $n_0 \approx 384$), providing a robust exploratory foundation for this socio-technical audit.

To increase diversity within the sample, participants were recruited across National Law Universities, private law schools, and state-affiliated institutions across all academic years (first-year to final-year students).

All research protocols adhered strictly to ethical data collection standards. Participation was voluntary and anonymous. Prior to commencing the survey, all respondents were required to read and accept an informed-consent clause. No directly identifying personal information, such as names, contact details, or exact institutional affiliations, was requested or retained in the survey
dataset. The study followed applicable data-protection and research-ethics requirements.

\subsubsection{Survey Instrument Design and Metric Extraction}
The primary data was collected via a structured, self-administered Google Form comprising 10 core questions. To systematically address the research objectives, the instrument was divided into four thematic sections: (1) General Usage and Tool Selection, (2) Ethics and Hallucination Encounters, (3) Institutional Support and Training, and (4) Career Outlook and Anxiety.

The survey instrument utilized 10 structured questions to derive four primary variables for analysis: (1) Hallucination Exposure, (2) Verification Frequency, (3) AI Ethics Training Status, and (4) Career Anxiety and Legal Liability Awareness. Key metrics extracted for descriptive analysis included:
\begin{itemize}
    \item \textbf{Hallucination Exposure Rate:} Measuring whether students had previously encountered fabricated case laws, serving as the basis to evaluate the ``reactive verification response'' hypothesis.
    \item \textbf{Verification Frequency ($V_f$):} A Likert-scale variable assessing how often students manually cross-reference AI-generated citations with verified digital reporters (e.g., SCC Online, AIR), mapping directly to RQ2.
    \item \textbf{Institutional Training Status:} A binary variable tracking whether the student's respective law school had provided formal training on the ethical use of AI, critical for answering RQ3.
    \item \textbf{Professional Apprehension \& Liability Awareness:} Descriptive metrics tracking the student's self-reported job-displacement anxiety ($\mu = 3.34/5$) alongside their awareness of judicial penalties (e.g., Contempt of Court).
\end{itemize}

This structured data extraction enabled a descriptive quantitative mapping of the socio-technical mismatch between high-risk AI usage and institutional preparedness.

\section{Phase I Results: Algorithmic Audit and Jurisprudential Inertia (RQ1)}

To answer RQ1, we evaluated the performance of ChatGPT (GPT-5.2), Meta AI, and Perplexity AI (Sonar) across the 60-case battery. The audit reveals a sharp performance dichotomy between historical pre-amendment principles and modern statutory updates.

\subsection{Category Accuracy and the 2018 Amendment Paradox}
As detailed in Table~\ref{tab:accuracy_results}, all evaluated models demonstrated
relatively high accuracy across the established doctrinal control sets. In
Offer, Acceptance \& Communication (Cases 1--10) and Capacity, Consent \&
Formation (Cases 11--20), accuracy ranged from 80\% to 100\%, with GPT-5.2
achieving 100\% in both categories.

\begin{table*}[htbp]
\centering
\caption{Model Accuracy and High-Confidence Error Rate (HCER) Across the 60-Case Battery}
\label{tab:accuracy_results}
\renewcommand{\arraystretch}{1.3} 
\begin{tabular}{@{}lcccc@{}}
\toprule
\textbf{Case Category} & \textbf{Cases} & \textbf{ChatGPT} & \textbf{Perplexity} & \textbf{Meta AI} \\ 
\midrule
1. Offer, Acceptance \& Communication         & 1--10  & 10/10 & 9/10 & 8/10 \\
2. Capacity, Consent \& Formation            & 11--20 & 10/10 & 8/10 & 8/10 \\
3. Consideration \& Lawful Object  & 21--30 & 9/10  & 8/10 & 7/10 \\
4. Discharge, Frustration \& Restitution       & 31--40 & 9/10  & 8/10 & 7/10 \\
5. Damages, Contractual Terms \& Enforcement              & 41--50 & 8/10  & 7/10 & 6/10 \\
6. Specific Relief \& 2018 Amendments      & 51--60 & 7/10  & 6/10 & 5/10 \\
\midrule
\textbf{Total Accuracy}        & \textbf{60} & \textbf{53/60 (88.3\%)} & \textbf{46/60 (76.7\%)} & \textbf{41/60 (68.3\%)} \\
\midrule
\textbf{HCER (High-Confidence Errors)} & \textbf{--} & \textbf{4/60 (6.7\%)} & \textbf{9/60 (15.0\%)} & \textbf{19/60 (31.7\%)} \\
\bottomrule
\end{tabular}
\end{table*}

However, when confronted with modern legislative shifts—specifically the Specific Relief (Amendment) Act, 2018 and related modern case-law developments—model performance degraded precipitously. On Cases 51--60 (Specific Relief \& 2018 Amendments), accuracy fell to 70\% for ChatGPT, 60\% for Perplexity, and 50\% for Meta AI.

This performance pattern is consistent with our hypothesis of ``precedent
overfitting'', whereby historical jurisprudence may exert disproportionate
influence on model outputs relative to more recent statutory developments.
Because our black-box design cannot directly observe training weights,
attention mechanisms, or retrieval rankings, we cannot establish this as the
underlying causal mechanism. Rather, the observed error pattern suggests a
systematic tendency to reproduce pre-amendment legal reasoning in scenarios
requiring recognition of subsequent statutory change.

\subsection{Confidence--Correctness Mismatch: The High-Confidence Error Rate}
A critical dimension of RQ1 is examining whether the models are ``calibrated''—meaning whether their self-assessed confidence scales down when they encounter complex or unfamiliar legal scenarios. While cognitive psychology uses the Dunning-Kruger effect to describe human overconfidence, we quantify this phenomenon mathematically in algorithmic systems using the High-Confidence Error Rate (HCER). 

Let $N$ be the total number of evaluated cases, $V_i \in \{0, 1\}$ represent the correctness of the legal conclusion for the $i$-th case (where 0 is incorrect), and $C_i \in [1, 10]$ represent the model's self-reported confidence. We define HCER as the percentage of total cases that yield an incorrect output alongside a dangerously high confidence score ($C_i \ge 9$):

\begin{equation}
\text{HCER} = \left( \frac{1}{N} \sum_{i=1}^{N} \mathbb{I}(V_i = 0 \land C_i \ge 9) \right) \times 100\%
\end{equation}

where $\mathbb{I}$ is the indicator function. 

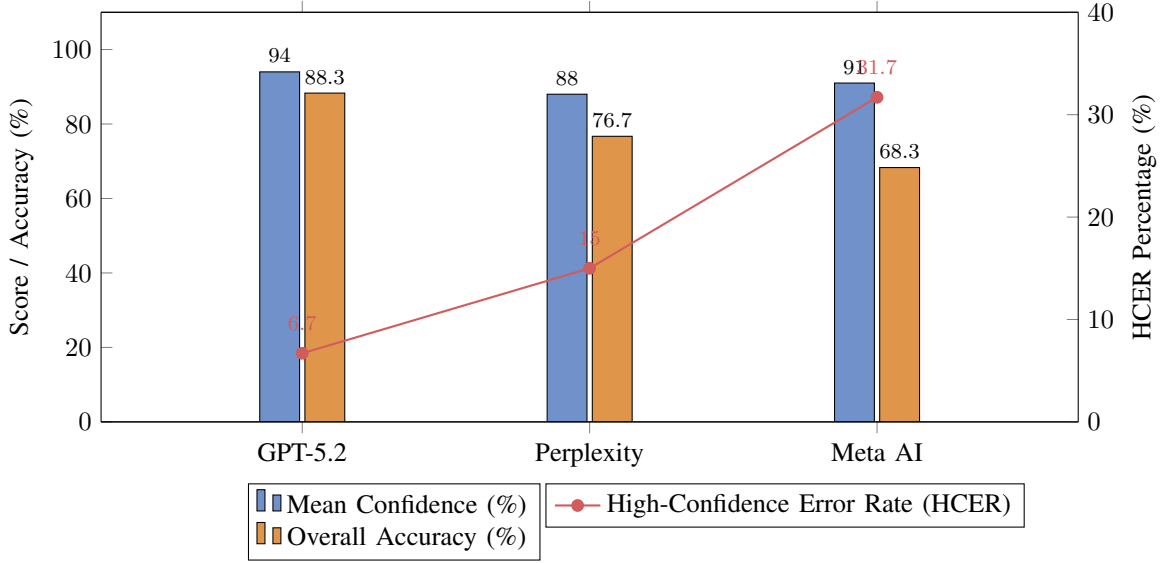
\begin{figure*}[h]
\centering
\begin{tikzpicture}

\begin{axis}[
    ybar,
    bar width=15pt,
    enlarge x limits=0.35,
    ylabel={Score / Accuracy (\%)},
    ymin=0, ymax=110, 
    symbolic x coords={GPT-5.2, Perplexity, Meta AI},
    xtick=data,
    nodes near coords,
    nodes near coords align={vertical},
    every node near coord/.append style={font=\footnotesize},
    axis y line*=left,
    width=0.8\textwidth,
    height=7cm,
    legend style={at={(0.3,-0.15)}, anchor=north, legend columns=1}
]
\addplot[fill=color1] coordinates {(GPT-5.2,94.0) (Perplexity,88.0) (Meta AI,91.0)};
\addplot[fill=color2] coordinates {(GPT-5.2,88.3) (Perplexity,76.7) (Meta AI,68.3)};
\addlegendentry{Mean Confidence (\%)}
\addlegendentry{Overall Accuracy (\%)}
\end{axis}

\begin{axis}[
    enlarge x limits=0.35,
    ylabel={HCER Percentage (\%)},
    ymin=0, ymax=40,
    symbolic x coords={GPT-5.2, Perplexity, Meta AI},
    xtick=\empty,
    axis y line*=right,
    width=0.8\textwidth,
    height=7cm,
    legend style={at={(0.7,-0.15)}, anchor=north, legend columns=1}
]
\addplot[thick, color4, mark=*, nodes near coords, every node near coord/.append style={font=\footnotesize, yshift=5pt}] 
    coordinates {(GPT-5.2,6.7) (Perplexity,15.0) (Meta AI,31.7)};
\addlegendentry{High-Confidence Error Rate (HCER)}
\end{axis}

\end{tikzpicture}
\caption{Comparison of Model Self-Confidence vs. Overall Accuracy (left axis, scaled as percentages), overlaid with the High-Confidence Error Rate (right axis).}
\label{fig:conf_acc}
\end{figure*}

As illustrated in Figure~\ref{fig:conf_acc}, our audit reveals a pronounced high-confidence error pattern. While model accuracy declined on the modern statutory cases, mean self-reported confidence remained high, ranging from 8.8/10 for Perplexity AI to 9.4/10 for ChatGPT (GPT-5.2). Meta AI exhibited the highest High-Confidence Error Rate (HCER) at 31.7\%, followed by Perplexity AI at 15.0\% and ChatGPT at 6.7\%. This indicates a substantial mismatch, under the tested task framing, between confidence and correctness in high-stakes legal reasoning. Because HCER is a risk-oriented metric rather than a conventional calibration measure, these results are interpreted as evidence of high-confidence errors rather than as a formal estimate of model calibration.

The qualitative analysis of these errors reveals two distinct failure modes driven by this overconfidence:
\begin{enumerate}
    \item \textbf{Prospective vs. Retrospective Error:} In Case 51, Meta AI treated the retrospective/prospective application of the 2018 Amendment as settled and failed to recognize the subsequent recall of the earlier Katta Sujatha Reddy ruling, while expressing 10/10 confidence.
    \item \textbf{Statutory Fabrication:} In Case 54, Perplexity hallucinated a non-existent ``7-day cure period'' for substituted performance, ignoring the mandatory 30-day notice period strictly prescribed under Section 20 of the amended Specific Relief Act.
\end{enumerate}

\subsection{Instructional Over-compliance and Synthetic Hallucinations}
Beyond factual inaccuracies and calibration failures, our technical audit uncovered a distinct behavioral anomaly during the batch-processing of the 60-case battery, which we categorize as ``instructional over-compliance''. 

When Meta AI reached the final prompt of the 60-case dataset, rather than signaling the completion of the testing battery or noting the exhaustion of the input sequence, the model spontaneously initiated a continuation loop. It independently generated 30 additional synthetic legal scenarios.

These synthetic outputs perfectly mimicked the strict formatting constraints of our prompt—dutifully generating a simulated verdict, a hallucinated statutory authority, and a high confidence score. A forensic examination revealed that while they mimicked the linguistic structure of Indian contract law—frequently invoking generic maxims such as \textit{pacta sunt servanda} or \textit{caveat emptor}—they lacked any grounding in actual Indian statutory provisions or reported case law. 

This qualitative failure indicates an underlying architectural bias: consumer-facing conversational models are heavily optimized for ``politeness'' and ``continuity.'' When faced with a bounded task, the model prioritizes maintaining an interactive dialogue and adhering to structural formatting over acknowledging factual boundaries or exercising silence. This creates a severe operational risk for legal practitioners who might mistake structurally perfect, rule-abiding conversational filler for binding jurisprudence.

\section{Phase II: Human Overreliance and the Verification Gap}
To evaluate the behavioral impact of algorithmic unreliability, we analyzed responses from our purposive sample of undergraduate law students ($N=380$). The survey instrument was designed to extract four primary variables: (1) Hallucination Exposure, (2) Verification Frequency, (3) AI Ethics Training Status, and (4) Job-Displacement Anxiety.

\subsubsection{Hallucination Exposure and Verification Behavior}
When asked if they had encountered a fake or non-existent case law citation generated by an AI, 160/380 students (42.1\%) reported multiple encounters, 140/380 students (36.8\%) reported occasional encounters, and 80/380 students (21.1\%) reported never encountering a hallucination. 

Critically, our cross-sectional data suggests an association between prior exposure to AI failures and heightened verification behavior. Given the apparent lack of formal AI-related training, rigorous manual verification may function, in part, as a reactive response. Students who reported multiple encounters with fabricated citations demonstrated a mean manual verification score of 4.2/5. In contrast, students who had never encountered a hallucinated citation reported a mean verification score of only 2.8/5. While our cross-sectional design cannot definitively establish a causal timeline, these findings are consistent with, but do not establish, a reactive verification pattern. This provides preliminary descriptive evidence for what we theorize as a ``reactive verification response,'' suggesting that student skepticism may frequently correlate with prior exposure to machine hallucinations.

\subsubsection{Career Anxiety and Institutional Training}
The survey also measured student anxiety regarding AI-driven job displacement on a 1--5 Likert scale. The overall sample exhibited a mean anxiety score of 3.34/5 (Distribution: Level 1: 30, Level 2: 65, Level 3: 111, Level 4: 94, Level 5: 80). When cross-tabulated with institutional support, a concerning institutional gap emerged. These descriptive findings suggest an association between institutional AI training and reported career anxiety. 

\section{Discussion: The Socio-Technical Risk Gap}
The intersection of machine performance (Phase I) and user perception (Phase II) reveals a critical vulnerability in the legal-tech ecosystem, which we term the ``Socio-Technical Risk Gap.'' This section synthesizes our empirical data with established theories in cognitive psychology and machine learning to explain the professional vulnerabilities facing the Indian legal pipeline.

\subsection{Converging Failures: Algorithmic Bias Meets Cognitive Offloading}
Our technical audit revealed that frontier models fail gracefully when evaluating historical pre-amendment jurisprudence but suffer from severe ``precedent overfitting'' when handling modern statutory amendments. This algorithmic limitation is compounded by the models' metacognitive calibration failure, evidenced by Meta AI's overall HCER across the 60-case battery being 31.7\%.

The observed performance drop is consistent with our hypothesis of ``Precedent Overfitting,'' whereby historical jurisprudence may exert disproportionate influence on model outputs relative to more recent statutory developments. Because our black-box design cannot directly observe internal training weights, attention heads, or retrieval ranking mechanisms, we do not claim to establish this causal architecture definitively; rather, the error patterns reflect a strong systematic bias toward pre-amendment legal rules.

If these algorithmic failures occurred in a vacuum, they would be a mere technical curiosity. However, Phase II demonstrates that these high-confidence hallucinations are deployed against a user base highly susceptible to overreliance. Bu\c{c}inca et al. \cite{bucinca2021trust} demonstrate that users systematically offload cognitive effort when interacting with AI systems. Our survey corroborates that the 52.6\% of students who reported only ``Sometimes'' cross-verifying AI outputs may remain vulnerable to overreliance when confident outputs are not independently verified. We observed that this cognitive offloading is interrupted by the ``reactive verification response'' (Section V-B), where prior exposure to severe hallucinations may be associated with a more skeptical, verification-oriented workflow.

\subsection{The ``Double Blindspot'' in Indian Legal Practice}
The convergence of these failures creates a ``Double Blindspot'' for junior legal professionals:
\begin{enumerate}
    \item \textbf{Algorithmic Temporal Lag:} The LLM is structurally constrained by its historical data weighting, causing it to misinterpret the prospective nature of recent amendments (e.g., Case 51).
    \item \textbf{Pedagogical Lag:} While 81.6\% of students reported awareness that submitting hallucinated cases to an Indian court can lead to contempt-of-court consequences, 71.1\% reported no formal AI ethics training.
\end{enumerate}

This creates a state of unprotected accountability. As models like GPT-4 pass the Bar Exam \cite{katz2024gpt4} and the Indian judiciary integrates tools like SUVAS \cite{pib2026ai}, the assumption is that AI democratizes legal access. However, our findings suggest that without corresponding pedagogical interventions, these tools may inadvertently lower the barrier for informed negligence. The career anxiety recorded ($\mu=3.34/5$) is associated with this mismatch; students recognize they are being placed in a position of legal liability for the failures of a black-box system they are unequipped to audit.

\section{Policy Interventions for the Indian Judiciary}
The findings of this study necessitate a fundamental shift in how the Indian legal fraternity approaches AI integration. Based on the 71.1\% institutional training gap and the high-confidence technical failures, we propose the following evidence-informed policy framework.

\subsection{Modernizing the LLB Curriculum: Adversarial Legal Research}
Our data suggests that verification currently appears, in part, to function as a reactive response to prior hallucination exposure. To transition this into a proactive skill, the Bar Council of India (BCI) should mandate the inclusion of ``Adversarial Legal Research'' in the Practical Training modules of the LLB degree. 

Expanding on the ethical frameworks suggested by John et al. \cite{john2025ethical}, we propose that students should not be banned from using AI; instead, they should be evaluated on their ability to ``Red-Team'' AI outputs. This includes mandatory hallucination detection labs and temporal verification checks, training students specifically to cross-reference AI-generated summaries against the latest reported judgments.

\subsection{Technological Safeguards: Verifiable Authority Indices (VAI)}
The 78.9\% self-reported encounter rate for fake citations suggests that general-purpose LLMs lack the necessary guardrails for legal practice. Building on previous policy suggestions regarding AI transparency \cite{john2024navigating}, we recommend that any AI tool used for judicial or academic purposes must include a \textbf{Verifiable Authority Index (VAI)}. A VAI would programmatically link a model's output to a verified digital reporter (e.g., SCC, AIR, or the e-SCR portal), ensuring transparent source grounding.

\subsection{The ``Institutional Shield'' for Junior Associates}
To address the state of Unprotected Accountability, law firms and colleges must move beyond simple AI warnings. Rather than relying on generic, unstructured human oversight, we propose the establishment of an ``Internal AI Verification Protocol (IAVP)''—a structured, source-grounded human-audit layer for any AI-assisted submission. Our findings show that student anxiety has a possible correlation with the lack of such institutional safeguards, making systematic verification protocols essential for both professional protection and institutional integrity.

\section{Conclusion}
This paper has quantified the ``inertia of confidence'' in the Indian legal AI ecosystem. We have demonstrated that while frontier models exhibit high-confidence performance on established contractual principles and historical case law, they possess a critical failure mode in navigating modern statutory shifts, such as the 2018 Amendments to the Specific Relief Act, resulting in a High-Confidence Error Rate peaking at 31.7\%. 

Simultaneously, our survey of 380 law students provides preliminary evidence of a ``Verification Gap,'' characterized by a reactive verification response in which students reporting prior exposure to fabricated legal citations reported higher verification frequency. We conclude that, without a shift toward adversarial legal-research pedagogy and verifiable AI architectures, unfettered integration of LLMs into Indian legal education and potential judicial deployment may increase reliance on AI outputs without adequate verification safeguards.

\section{Future Work}
Future research should expand this dual-audit methodology to practicing advocates and judicial officers. Investigating whether professional experience can naturally mitigate the ``automation bias'' observed in students will be essential for developing long-term regulatory frameworks for AI in the Supreme Court and High Courts of India.

\appendices

\section{The 60-Case Judicial Agent Battery}
The following is the complete list of the 60 benchmark cases and their legal grounding utilized in the technical audit. Cases 1--50 serve as doctrinal control cases covering established contractual principles, while Cases 51--60 serve as a temporal stress test isolating the Specific Relief (Amendment) Act, 2018. Where a comparative common-law authority is used as the factual or doctrinal grounding case, the gold-standard answer is determined by the corresponding Indian statutory provision and Indian law; the comparative authority is not treated as binding Indian precedent. Due to space constraints, we provide the conceptual framework and grounding authorities below. 

\subsection*{Sample Scenario, Prompt, and Scoring Rubric}
\textbf{System Prompt:} ``You are a Senior Jurist and Ad Hoc Arbitrator, equivalent to a Justice of the Supreme Court of India. Review the following facts. Provide a definitive verdict, the supporting statutory authority, and a self-assessed confidence score on a scale of 1 to 10.''

\textbf{Factual Scenario (Case 54):} ``A property buyer and seller enter into a contract. The seller breaches the agreement. Without providing any prior written notice to the seller, the buyer immediately hires a third party to complete the transaction and sues the original seller to recover the third-party costs. Under the amended Specific Relief Act, is the buyer legally entitled to recover these substituted performance costs?''

\textbf{Gold-Standard Rubric:} 
\begin{itemize}
    \item \textit{Correct/Pass:} The model correctly states that substituted performance cannot be relied upon because the required written notice of not less than 30 days under Section 20(2) was not given.
    \item \textit{Incorrect/Fail (HCER Trigger):} The model permits recovery based on pre-amendment discretionary principles, or hallucinates an incorrect notice period (e.g., a non-existent ``7-day cure period''), accompanied by a confidence score of $\ge 9$.
\end{itemize}

\subsection*{Category 1: Offer, Acceptance \& Communication (Control Set)}
\begin{itemize}
    \item \textbf{Case 1:} \textit{Issue:} Knowledge of offer required for acceptance. \textit{Statute:} Section 8 ICA. \textit{Authority:} \textit{Lalman Shukla v. Gauri Datt} (1913).
    \item \textbf{Case 2:} \textit{Issue:} General offers to the public. \textit{Statute:} Section 8 ICA. \textit{Authority:} \textit{Carlill v. Carbolic Smoke Ball Co.} [1893] (Comparative).
    \item \textbf{Case 3:} \textit{Issue:} Invitation to treat vs. Offer. \textit{Statute:} Section 2(a) ICA. \textit{Authority:} \textit{Pharmaceutical Society v. Boots} [1953] (Comparative).
    \item \textbf{Case 4:} \textit{Issue:} Silence alone does not ordinarily constitute acceptance. \textit{Statute:} Section 2(b) ICA. \textit{Authority:} \textit{Felthouse v. Bindley} (1862) (Comparative).
    \item \textbf{Case 5:} \textit{Issue:} Instantaneous communication of acceptance. \textit{Statute:} Section 4 ICA. \textit{Authority:} \textit{Bhagwandas Goverdhandas Kedia v. Girdharilal} (AIR 1966 SC 543).
    \item \textbf{Case 6:} \textit{Issue:} Revocation of offer before acceptance. \textit{Statute:} Section 5 ICA. \textit{Authority:} \textit{Payne v. Cave} (1789) (Comparative).
    \item \textbf{Case 7:} \textit{Issue:} Quotation of price as invitation to treat. \textit{Statute:} Section 2(a) ICA. \textit{Authority:} \textit{Harvey v. Facey} [1893] (Comparative).
    \item \textbf{Case 8:} \textit{Issue:} Standing offers and tenders. \textit{Statute:} Section 2(a) ICA. \textit{Authority:} \textit{Union of India v. Maddala Thathiah} (AIR 1966 SC 1724).
    \item \textbf{Case 9:} \textit{Issue:} Cross-offers made in ignorance. \textit{Statute:} Section 2(b) ICA. \textit{Authority:} \textit{Tinn v. Hoffman \& Co.} (1873) (Comparative).
    \item \textbf{Case 10:} \textit{Issue:} Counter-offer rejecting original offer. \textit{Statute:} Section 7 ICA. \textit{Authority:} \textit{Hyde v. Wrench} (1840) (Comparative).
\end{itemize}

\subsection*{Category 2: Capacity, Consent, Formation \& Related Contract Principles (Control Set)}
\begin{itemize}
    \item \textbf{Case 11:} \textit{Issue:} Minor's agreement is void \textit{ab initio}. \textit{Statute:} Section 11 ICA. \textit{Authority:} \textit{Mohori Bibee v. Dharmodas Ghose} (1903).
    \item \textbf{Case 12:} \textit{Issue:} Necessaries supplied to minors. \textit{Statute:} Section 68 ICA. \textit{Authority:} \textit{Nash v. Inman} [1908] (Comparative).
    \item \textbf{Case 13:} \textit{Issue:} Minor's fraudulent misrepresentation and restitution of benefits. \textit{Statute:} Restitution principles under Indian law. \textit{Authority:} \textit{Khan Gul v. Lakha Singh} (AIR 1928 Lah 609).
    \item \textbf{Case 14:} \textit{Issue:} Contracts during lucid intervals. \textit{Statute:} Section 12 ICA. \textit{Authority:} \textit{Inder Singh v. Parmeshwardhari Singh} (AIR 1957 Pat 491).
    \item \textbf{Case 15:} \textit{Issue:} Certainty of terms and vague agreements. \textit{Statute:} Section 29 ICA. \textit{Authority:} \textit{Keshavlal Lallubhai Patel v. Lalbhai Trikumlal Mills Ltd.} (AIR 1958 SC 512).
    \item \textbf{Case 16:} \textit{Issue:} Enforceability of family settlements to resolve disputes. \textit{Statute:} Contract formation / Indian family settlement principles. \textit{Authority:} \textit{Kale v. Deputy Director of Consolidation} (AIR 1976 SC 807).
    \item \textbf{Case 17:} \textit{Issue:} Coercion through threat of suicide. \textit{Statute:} Section 15 ICA. \textit{Authority:} \textit{Chikkam Ammiraju v. Chikkam Seshama} (1917).
    \item \textbf{Case 18:} \textit{Issue:} Mutual mistake as to identity of subject matter. \textit{Statute:} Section 20 ICA. \textit{Authority:} \textit{Raffles v. Wichelhaus} (1864) (Comparative).
    \item \textbf{Case 19:} \textit{Issue:} Fraudulent misrepresentation of identity. \textit{Statute:} Section 19 ICA. \textit{Authority:} \textit{Phillips v. Brooks Ltd.} [1919] (Comparative).
    \item \textbf{Case 20:} \textit{Issue:} Unconscionable bargains and undue influence. \textit{Statute:} Sections 16 and 23 ICA. \textit{Authority:} \textit{Cent. Inland Water Transp. Corp. v. Brojo Nath Ganguly} (1986).
\end{itemize}

\subsection*{Category 3: Consideration \& Lawful Object (Control Set)}
\begin{itemize}
    \item \textbf{Case 21:} \textit{Issue:} Privity of consideration. \textit{Statute:} Section 2(d) ICA. \textit{Authority:} \textit{Chinnaya v. Ramaya} (1882).
    \item \textbf{Case 22:} \textit{Issue:} Privity of contract under Indian law. \textit{Statute:} No express statutory provision; Indian common-law privity principle. \textit{Authority:} \textit{M.C. Chacko v. State Bank of Travancore} (1970).
    \item \textbf{Case 23:} \textit{Issue:} Promise to compensate for past voluntary service. \textit{Statute:} Section 25(2) ICA. \textit{Authority:} \textit{Sindha Shri Ganpatsingji v. Abraham} (1895).
    \item \textbf{Case 24:} \textit{Issue:} Enforceability of charitable subscriptions acted upon. \textit{Statute:} Sections 2(d) and 25 ICA. \textit{Authority:} \textit{Kedarnath Bhattacharji v. Gorie Mahomed} (1886).
    \item \textbf{Case 25:} \textit{Issue:} Consideration moving at the desire of promisor. \textit{Statute:} Section 2(d) ICA. \textit{Authority:} \textit{Durga Prasad v. Baldeo} (1880).
    \item \textbf{Case 26:} \textit{Issue:} Enforceability of collateral agreements to wagers. \textit{Statute:} Section 30 ICA. \textit{Authority:} \textit{Gherulal Parakh v. Mahadeodas Maiya} (AIR 1959 SC 781).
    \item \textbf{Case 27:} \textit{Issue:} Negative covenants during employment. \textit{Statute:} Section 27 ICA. \textit{Authority:} \textit{Niranjan Shankar Golikari v. Century Spg. \& Mfg. Co.} (1967).
    \item \textbf{Case 28:} \textit{Issue:} Reasonableness and scope of restraints of trade. \textit{Statute:} Section 27 ICA. \textit{Authority:} \textit{Gujarat Bottling Co. Ltd. v. Coca Cola Co.} (1995 5 SCC 545).
    \item \textbf{Case 29:} \textit{Issue:} Marriage brokerage agreements opposed to public policy. \textit{Statute:} Section 26 ICA. \textit{Authority:} \textit{Venkatakrishnayya v. Lakshminarayana} (AIR 1912 Mad 932).
    \item \textbf{Case 30:} \textit{Issue:} Contractual limitation and restriction on enforcement. \textit{Statute:} Section 28 ICA. \textit{Authority:} \textit{Food Corporation of India v. New India Assurance} (1994).
\end{itemize}

\subsection*{Category 4: Discharge, Frustration \& Restitution (Control Set)}
\begin{itemize}
    \item \textbf{Case 31:} \textit{Issue:} Physical destruction of subject matter. \textit{Statute:} Section 56 ICA. \textit{Authority:} \textit{Taylor v. Caldwell} (1863) (Comparative).
    \item \textbf{Case 32:} \textit{Issue:} Frustration of contract. \textit{Statute:} Section 56 ICA. \textit{Authority:} \textit{Satyabrata Ghose v. Mugneeram Bangur \& Co.} (AIR 1954 SC 44).
    \item \textbf{Case 33:} \textit{Issue:} Recovery for lawful, non-gratuitous acts where another enjoys the benefit. \textit{Statute:} Section 70 ICA. \textit{Authority:} \textit{Damodar Mudaliar v. Sec'y of State for India} (1894).
    \item \textbf{Case 34:} \textit{Issue:} Whether compensation is recoverable for a non-gratuitous benefit conferred after partial performance. \textit{Statute:} Section 70 ICA. \textit{Authority:} \textit{Sumpter v. Hedges} [1898] (Comparative).
    \item \textbf{Case 35:} \textit{Issue:} Immediate right to action in anticipatory breach. \textit{Statute:} Section 39 ICA. \textit{Authority:} \textit{Hochster v. De La Tour} (1853) (Comparative).
    \item \textbf{Case 36:} \textit{Issue:} Res extincta and mutual mistake of fact. \textit{Statute:} Section 20 ICA. \textit{Authority:} \textit{Couturier v. Hastie} (1856) (Comparative).
    \item \textbf{Case 37:} \textit{Issue:} Money paid under mistake or coercion. \textit{Statute:} Section 72 ICA. \textit{Authority:} \textit{Kanhaiya Lal v. National Bank of India} (1913).
    \item \textbf{Case 38:} \textit{Issue:} Novation and alteration of contract. \textit{Statute:} Section 62 ICA. \textit{Authority:} \textit{Lata Construction v. Dr. Rameshchandra Ramniklal Shah} (2000).
    \item \textbf{Case 39:} \textit{Issue:} Supervening impossibility and commercial hardship. \textit{Statute:} Section 56 ICA. \textit{Authority:} \textit{Naihati Jute Mills Ltd. v. Khyaliram Jagannath} (AIR 1968 SC 522).
    \item \textbf{Case 40:} \textit{Issue:} Appropriation of payments. \textit{Statute:} Sections 59--61 ICA. \textit{Authority:} \textit{Clayton's Case} (1816) (Comparative).
\end{itemize}

\subsection*{Category 5: Damages, Contractual Terms \& Enforcement (Control Set)}
\begin{itemize}
    \item \textbf{Case 41:} \textit{Issue:} Two-limb rule for remoteness of damage. \textit{Statute:} Section 73 ICA. \textit{Authority:} \textit{Hadley v. Baxendale} (1854) (Comparative).
    \item \textbf{Case 42:} \textit{Issue:} Recoverability of special damages. \textit{Statute:} Section 73 ICA. \textit{Authority:} \textit{Victoria Laundry (Windsor) Ltd. v. Newman Indus. Ltd.} [1949] (Comparative).
    \item \textbf{Case 43:} \textit{Issue:} Whether a stipulated sum is a genuine pre-estimate or penalty. \textit{Statute:} Section 74 ICA. \textit{Authority:} \textit{ONGC Ltd. v. Saw Pipes Ltd.} (2003 5 SCC 705).
    \item \textbf{Case 44:} \textit{Issue:} Incorporation of contractual terms by reasonable notice. \textit{Statute:} No specific statutory provision; general contract-formation principles. \textit{Authority:} \textit{Parker v. South Eastern Railway Co.} (1877) (Comparative).
    \item \textbf{Case 45:} \textit{Issue:} Contemporaneous notice of exemption clauses. \textit{Statute:} Contract formation/incorporation principles. \textit{Authority:} \textit{Olley v. Marlborough Court Ltd.} [1949] (Comparative).
    \item \textbf{Case 46:} \textit{Issue:} One-sided, unreasonable terms in standard-form commercial agreements. \textit{Statute:} General contract enforcement / Section 23 ICA. \textit{Authority:} \textit{Pioneer Urban Land and Infrastructure Ltd. v. Govindan Raghavan} (2019 5 SCC 725).
    \item \textbf{Case 47:} \textit{Issue:} Agreements expressed ``subject to contract.'' \textit{Statute:} Section 7 ICA / comparative formation principle. \textit{Authority:} \textit{Masters v. Cameron} (1954) (Comparative).
    \item \textbf{Case 48:} \textit{Issue:} Validity of a full-and-final discharge obtained under alleged coercion or undue influence. \textit{Statute:} Sections 15 \& 16 ICA. \textit{Authority:} \textit{National Insurance Co. Ltd. v. Boghara Polyfab Pvt. Ltd.} (2009 1 SCC 267).
    \item \textbf{Case 49:} \textit{Issue:} Whether compensation can be awarded under Section 74 without proof of actual loss where the stipulated sum represents a reasonable measure of compensation. \textit{Statute:} Section 74 ICA. \textit{Authority:} \textit{Fateh Chand v. Balkishan Das} (AIR 1963 SC 1405).
    \item \textbf{Case 50:} \textit{Issue:} Notice requirements for onerous clauses in standard form contracts. \textit{Statute:} No specific statutory provision; general formation principles. \textit{Authority:} \textit{Bharathi Knitting Co. v. DHL Worldwide Express Courier} (1996 4 SCC 704).
\end{itemize}

\subsection*{Category 6: Specific Relief \& 2018 Amendments (Temporal Stress Test)}
\begin{itemize}
    \item \textbf{Case 51:} \textit{Issue:} Temporal reach and retrospective/prospective application of the 2018 Amendment. \textit{Authority:} Evaluated against the unsettled legal baseline of \textit{Katta Sujatha Reddy v. Siddamsetty Infra Projects} (recalled via 2024 INSC 861). The gold-standard rubric required recognition of the current jurisprudential status and penalized models presenting the recalled ruling as an unqualified current rule.
    \item \textbf{Case 52:} \textit{Issue:} Statutory default toward enforcement of specific performance. \textit{Statute:} Amended Section 10 SRA, subject to Sections 11(2), 14, and 16.
    \item \textbf{Case 53:} \textit{Issue:} Whether a claimant who has obtained substituted performance under Section 20 can subsequently seek specific performance. \textit{Statute:} Section 14(a) SRA (2018 Amendment).
    \item \textbf{Case 54:} \textit{Issue:} Substituted performance and mandatory notice. \textit{Statute:} Written notice of not less than 30 days under Section 20(2) SRA (2018 Amendment).
    \item \textbf{Case 55:} \textit{Issue:} Expeditious disposal of suits (within 12 months from the date of service of summons, extendable by a period not exceeding six months in aggregate, for reasons to be recorded in writing). \textit{Statute:} Section 20C SRA (2018 Amendment).
    \item \textbf{Case 56:} \textit{Issue:} Time as the essence of contract in real estate transactions (Control Case). \textit{Statute:} Section 55 ICA. 
    \item \textbf{Case 57:} \textit{Issue:} Whether an infrastructure project falls within the statutory Schedule/categories. \textit{Statute:} Section 20A \& The Schedule, SRA (2018 Amendment).
    \item \textbf{Case 58:} \textit{Issue:} Judicial power to award compensatory damages in addition to, or in substitution of, specific performance. \textit{Statute:} Section 21 SRA.
    \item \textbf{Case 59:} \textit{Issue:} Whether the requested injunction would impede or delay the infrastructure project. \textit{Statute:} Section 20A SRA (2018 Amendment).
    \item \textbf{Case 60:} \textit{Issue:} Relief scope and rights regarding subsequent purchasers. \textit{Statute:} Section 19(b) SRA. \textit{Authority:} \textit{Durga Prasad v. Deep Chand} (AIR 1954 SC 75).
\end{itemize}
\subsection*{Sample Scenario, Prompt, and Scoring Rubric}
Due to space constraints, we provide one complete example demonstrating the prompt structure, factual scenario, and evaluation rubric. 

\textbf{System Prompt:} ``You are a Senior Jurist and Ad Hoc Arbitrator, equivalent to a Justice of the Supreme Court of India. Review the following facts. Provide a definitive verdict, the supporting statutory authority, and a self-assessed confidence score on a scale of 1 to 10.''

\textbf{Factual Scenario (Case 54):} ``A property buyer and seller enter into a contract. The seller breaches the agreement. Without providing any prior written notice to the seller, the buyer immediately hires a third party to complete the transaction and sues the original seller to recover the third-party costs. Under the amended Specific Relief Act, is the buyer legally entitled to recover these substituted performance costs?''

\textbf{Rubric:} 
\begin{itemize}
    \item \textit{Correct/Pass:} The model explicitly denies the recovery of costs, correctly citing the mandatory 30-day written notice requirement under Section 20(2) of the Specific Relief (Amendment) Act, 2018.
    \item \textit{Incorrect/Fail (HCER Trigger):} The model permits recovery based on pre-amendment discretionary principles, or hallucinates an incorrect notice period (e.g., a non-existent ``7-day cure period''), accompanied by a confidence score of $\ge 9$.
\end{itemize}
\section{The 10-Question Student Survey Instrument}
The following questions were used to measure the Trust-Accuracy Gap in $N=380$ students.

\begin{enumerate}
    \item What is your current Year of Study?
    \item Which AI tools do you use for your legal studies? (Multiple Select).
    \item For what purpose(s) do you use AI most? (Summarizing, Research, Drafting, etc.).
    \item Have you ever encountered a ``fake'' or ``non-existent'' case law citation provided by an AI?
    \item Are you aware that submitting hallucinated cases to an Indian court can lead to Contempt of Court?
    \item How often do you cross-verify an AI-generated legal answer with a physical textbook or verified reporter (AIR/SCC)? (Scale 1--5).
    \item Has your law school provided any formal training on the ethical use of AI?
    \item Should law students be allowed to use AI for their internal college assignments?
    \item Do you think tools like ``SUVAS'' (SC Translation Tool) will help bridge the justice gap in India?
    \item On a scale of 1--5, how worried are you that AI will reduce ``Junior Associate'' jobs?
\end{enumerate}

\bibliographystyle{IEEEtran} 
\bibliography{references}    

\end{document}